\documentclass{article}

\usepackage{PRIMEarxiv}

\usepackage[utf8]{inputenc} 
\usepackage[T1]{fontenc}    
\usepackage{hyperref}       
\usepackage{url}            
\usepackage{booktabs}       
\usepackage{amsfonts}       
\usepackage{nicefrac}       
\usepackage{microtype}      
\usepackage{lipsum}
\usepackage{fancyhdr}       
\usepackage{graphicx}       
\graphicspath{{media/}}     
\usepackage{multirow}
\usepackage{subcaption}
\usepackage{siunitx}
\usepackage[most]{tcolorbox}
\usepackage{xcolor} 
\usepackage{natbib}
\tcbset{
  promptbox/.style={
    colback=gray!10,    
    colframe=gray!80,   
    fonttitle=\bfseries,
    title=LLM Prompt,   
    boxrule=0.5pt,      
    arc=3pt,            
    outer arc=3pt,
    top=4pt, bottom=4pt,
    left=6pt, right=6pt
  }
}

\title{Exploring LLM-based Frameworks for Fault Diagnosis
}

\author{
  Xian Yeow Lee, Lasitha Vidyaratne, Ahmed Farahat, Chetan Gupta \\
  Hitachi America Ltd., R\&D, \\
  Santa Clara \\
  \texttt{\{xian.lee, lasitha.vidyaratne, ahmed.farahat, chetan.gupta\}@hal.hitachi.com} \\
}

\begin{document}
\maketitle

\begin{abstract}
Large Language Model (LLM)-based systems present new opportunities for autonomous health monitoring in sensor-rich industrial environments. This study explores the potential of LLMs to detect and classify faults directly from sensor data, while producing inherently explainable outputs through natural language reasoning. We systematically evaluate how LLM-system architecture (single-LLM vs. multi-LLM), input representations (raw vs. descriptive statistics), and context window size affect diagnostic performance. Our findings show that LLM systems perform most effectively when provided with summarized statistical inputs, and that systems with multiple LLMs using specialized prompts offer improved sensitivity for fault classification compared to single-LLM systems. While LLMs can produce detailed and human-readable justifications for their decisions, we observe limitations in their ability to adapt over time in continual learning settings, often struggling to calibrate predictions during repeated fault cycles. These insights point to both the promise and the current boundaries of LLM-based systems as transparent, adaptive diagnostic tools in complex environments.
\end{abstract}

\keywords{Agents, Large language models, Fault diagnosis}

\section{Introduction}

Recent advances in the capabilities of large language models (LLMs) have significantly enhanced the practicality of artificial intelligence, particularly for tasks involving language understanding, reasoning, and decision-making. Trained on vast textual corpora, LLMs have evolved from simple text-completion tools into general-purpose agents capable of planning, tool use, memory retrieval, and multi-step reasoning~\cite{fu2023specializing,li2025review}. In this context, LLM-based systems have emerged as tools that leverage language models to perform complex reasoning, provide explanations, and interact via natural language interfaces~\cite{openai_agents}.

In the field of Prognostics and Health Management (PHM), there is a growing demand for intelligent systems that go beyond fault detection and diagnosis to integrate seamlessly with human workflows. LLM-based systems address this need by providing intuitive interfaces between data-driven diagnostics and domain experts. Because they operate through natural language, these systems reduce the need for extensive operational expertise in their design and usage. Unlike traditional black-box models, LLM-powered systems can explain their inferences, seek clarifications from users, and adapt to changing operational contexts through in-context learning.

Although LLM-based systems has demonstrated utility in areas such as document summarization, workflow automation, and decision support across domains like finance, healthcare, and software development~\cite{10849561}, its application in engineering fields such as PHM remains underexplored. These domains typically involve high-frequency numerical sensor data rather than structured text or prompts. A central question is whether LLMs can be adapted to operate effectively on such data and reliably detect anomalies and diagnose faults in complex physical systems.

By studying LLM-based systems in a sensor-driven domain, we contribute to a growing body of research seeking to bridge the gap between general-purpose AI and domain-specific engineering applications. Our findings suggest that while LLM-based systems shows strong potential in PHM, important challenges remain for enabling real-time, robust reasoning in dynamic environments. Our contributions are as follows:
\begin{itemize}
\item We develop a simulated HVAC system with multiple sensors and configurable fault types, allowing controlled experimentation across diverse fault scenarios and conditions.
\item We propose an LLM-based framework for anomaly and fault detection, featuring multi-representation data interpretation, in-context learning, and natural language explanations.~\footnote{Simulator, code and data available at \\ ~\text{https://github.com/xylhal/PHM\_LLMFaultDiagnosis}}
\item We conduct comprehensive evaluation across different data representations (raw, statistical, and hybrid), time-windowing strategies, and performance metrics for anomaly and fault detection, including the system's capacity for continual learning.
\end{itemize}

\section{Related works}

\noindent \textbf{Traditional Methods} \\
Traditional anomaly and fault detection in time-series sensor data relies on statistical and signal-processing methods~(~\cite{rousseeuw2018anomaly}). Early techniques include threshold-based control charts and statistical tests to identify outliers~(~\cite{aerospace6110117}). Subspace methods like Principal Component Analysis (PCA) monitor deviations in sensor covariance, often after preprocessing such as multi-resolution wavelet transforms, as proposed by~\citet{jiang2017probabilistic}. Meanwhile, distance and density-based methods (e.g., k-NN, Local Outlier Factor) treat points in low-density regions as anomalies but often struggle with high-dimensional, large-scale data(~\cite{aerospace6110117}).

\noindent \textbf{Machine Learning Approaches} \\
With the rise of data-driven methods, machine learning techniques have gained traction in the PHM domain by learning normal system behavior and detecting deviations(~\cite{su2023machine}). For instance, \citet{rocchetta2022robust} combine multiple SVMs with system-structure models to detect component failures, optimizing hyperparameters to minimize system-level errors under limited failure data. Deep learning approaches, such as LSTM-based and variational autoencoders, detect anomalies in sensor data via reconstruction errors(~\cite{zhang2019review}). For example, \citet{liang2024anomaly} apply Transformer autoencoders for unsupervised anomaly detection, using residual-based statistical tests to flag faults. Other studies use convolutional networks or deep regression to identify anomalous patterns directly from raw data(~\cite{10194112}). While these models perform well on complex datasets, they often sacrifice interpretability and typically require substantial labeled data. To address this, unsupervised techniques such as DBSCAN and k-means cluster normal behavior and flag deviations(~\cite{usmani2022review}), while ensemble methods such as \citet{vanem2021unsupervised}'s work on marine engine sensor clustering combine multiple models to improve robustness. Despite their potential, ML-based approaches require careful tuning and validation to avoid overfitting and ensure reliable performance.

\noindent \textbf{LLM-Based Methods} \\
Recent work has explored the use of LLMs for time-series tasks by encoding signals as token sequences. \citet{alnegheimish2024can} propose SigLLM, which uses LLMs to label anomalies via prompting or detect them through forecasting by comparing predictions to actual signals. The forecasting approach outperforms prompting and a moving-average baseline, though specialized neural models still yield higher scores. \citet{jin2023time} introduce Time-LLM, which adapts frozen LLMs for forecasting using a prompt-as-prefix strategy. While it performs well in few-shot settings, later studies show that simpler attention layers can achieve comparable or better results. More recently,~\citet{gu2025argos} present Argos, an LLM-based anomaly detector for cloud infrastructure metrics. Argos uses multiple LLM-powered agents to autonomously generate, verify, and refine interpretable rule-based detectors, achieving strong performance on both public and industrial datasets. 

In this work, we build on these ideas but specifically study the ability of LLMs to detect multiple simultaneous faults in data derived from a simulated engineered HVAC system, comparing performance across different input representations.

\section{HVAC Simulation Environment}
To study the LLM agent's ability to detect faults from realistic sensor data, we developed a simulator that emulates the behavior of commercial HVAC systems under different operational conditions. The simulator models key components such as the compressor, heat exchanger, and air handling units; producing multi-sensor time-series data. The environment includes ambient conditions, thermal dynamics, power consumption, and supports fault injection with multi-sensor effects as well as capability to include sensor drifts.

\subsection{Thermal and Physical Models}

\subsubsection{Temperature Dynamics}
We begin with  the model of the temperature dynamics, where the indoor temperature is governed by a simplified first-order thermal model:
\\
{
\begin{equation}
    T_{in}(t{+}1) = T_{in}(t) + \alpha(T_{amb}(t) {-} T_{in}(t)) - \beta Q_{cool}(t)
\end{equation}
}
\\
where $T_{in}$ is the indoor temperature, $T_{amb}$ is the ambient temperature, $Q_{cool}$ is cooling output, $\alpha$ is the thermal gain coefficient, and $\beta$ is cooling effectiveness. Ambient temperature is modeled as a daily sinusoidal cycle with noise:
\\
{\begin{equation}
    T_{amb}(t) = T_{mean} + A \sin\left(2\pi(t{-}\phi)/24\right) + \epsilon(t)
\end{equation}
}
\\
where $T_{mean}$ is the mean temperature, $A$ is amplitude, $\phi$ is the phase offset, and $\epsilon(t) \sim \mathcal{N}(0, \sigma^2)$.

\subsubsection{Pressure and Power Models}
To model the mechanical part of the system, we assume the suction $P_{suct}(t)$ and discharge $P_{disc}(t)$ pressures, i.e., pressure of refrigerant entering and exiting the compressor, scale with the normalized cooling demand, where $Q_{nom}$ denotes the nominal cooling capacity of the system:
\\
{
\begin{equation}
    P_{suct}(t) = P_0 + \gamma_1 \frac{Q_{cool}(t)}{Q_{nom}}, \quad
    P_{disc}(t) = P_0 + \gamma_2 \frac{Q_{cool}(t)}{Q_{nom}}
\end{equation}
}
\\
and $P_0$ denotes base pressure, and $\gamma_1$, $\gamma_2$ are scaling factors. Furthermore, the compressor power is modeled as:
\\
{
\begin{equation}
    P_{comp}(t) = 
    \begin{cases}
        P_{nom} + \eta(t) & \text{if cooling is active} \\
        0 & \text{otherwise}
    \end{cases}
\end{equation}
}
\\
where $P_{nom}$ is the nominal power consumption of the compressor and $\eta(t) \sim \mathcal{N}(0, \sigma_p^2)$ representing power variation.

\subsection{Fault Modeling and Sensor Effects}

To model potential faults in the system, we assume the fault dynamics can be parameterized by their severity $S$ and different temporal functions $f(t)$:
\\
{
\begin{equation}
    F(t) = S \cdot f(t)
\end{equation}
}
\\
We consider the following common onset profiles :
\begin{itemize}
  \item \textbf{Step:} $f(t) = 1$
  \item \textbf{Linear ramp:} $f(t) = \min\left(1, \frac{t - t_0}{t_1 - t_0} \right)$
  \item \textbf{Exponential:} $f(t) = 1 - e^{-\frac{t - t_0}{\tau}}$
\end{itemize}
where $t_0$ and $t_1$ denote the fault onset and saturation times and $\tau$ is defined as the exponential time constant

In our work, we consider three types of potential faults: refrigerant leaks, compressor faults and filter blockage. The faults are modeled to simultaneously influence several system variables, resulting in correlated patterns in the sensor data. Given that $T_{supply}$, $T_{return}$, and $Q_{air}$ denote the supply air temperature, return air temperature, and airflow rate respectively, and the other notations are as defined above, we define:

\textbf{Refrigerant Leak ($F_{leak}$):}
{
\begin{align}
    Q_{cool} &= Q_{cool}(1 - 0.5 F_{leak}) \\
    P_{suct} &= P_{suct}(1 - 0.3 F_{leak}) \\
    P_{comp} &= P_{comp}(1 + 0.2 F_{leak}) \\
    T_{supply} &= T_{supply} + 3.0 F_{leak}
\end{align}
}
\textbf{Compressor Fault ($F_{comp}$):}
{
\begin{align}
    P_{comp} &= P_{comp}(1 - 0.7 F_{comp}) \\
    Q_{cool} &= Q_{cool}(1 - 0.9 F_{comp}) \\
    P_{suct} &= P_{suct}(1 + 0.2 F_{comp}) \\
    P_{disc} &= P_{disc}(1 - 0.5 F_{comp})
\end{align}
}
\textbf{Filter Blockage ($F_{filter}$):}
{
\begin{align}
    Q_{air} &= Q_{air}(1 - 0.4 F_{filter}) \\
    T_{return} &= T_{return} + 2.0 F_{filter} \\
    P_{comp} &= P_{comp}(1 + 0.15 F_{filter})
\end{align}
}

In summary, the simulator takes in a set of user configurations such as ambient mean temperature, target indoor temperature, and cooling capacity, simulates the dynamics of the system, and generates time-series data for the following nine sensors: $T_{amb}$, $T_{in}$, $P_{comp}$, $Q_{cool}$, $P_{suct}$, $P_{disc}$, $T_{supply}$, $T_{return}$, and $Q_{air}$. This data is then used by the LLM agent to detect anomalies and faults.


\section{Experimental Method}

\subsection{Overview}

\begin{figure}[h]
\centering
\includegraphics[width=0.6\linewidth]{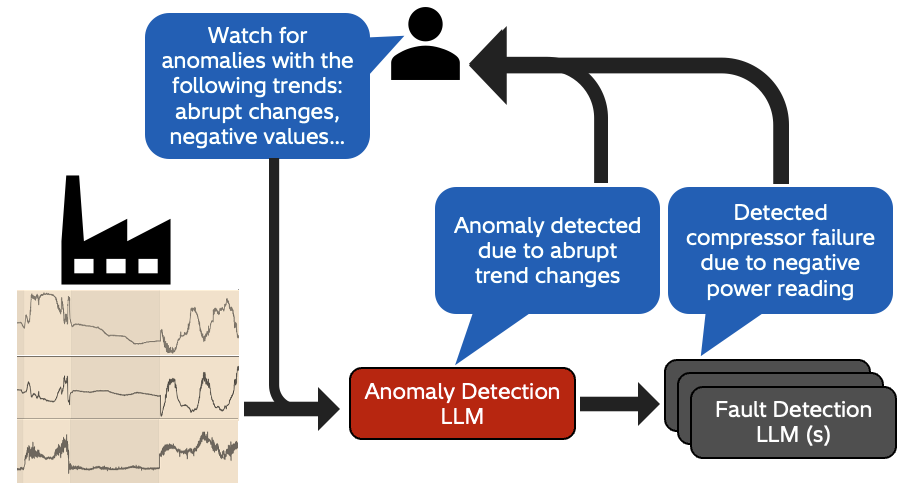}
\caption{Overview of an LLM-based framework for fault diagnosis. LLM-based capabilities enables operator to describe faults in natural language and predictions made by the framework are explainable.}
\label{fig:overview}
\end{figure}

This section outlines the experimental setup used to evaluate the capabilities of LLM-based system in the context of fault detection. We formulate the task as a multi-stage problem involving two primary LLMs: an anomaly detection LLM and possibly multiple classification LLM(s), shown in Figure~\ref{fig:overview}

In the first stage, the anomaly detection LLM receives time-series sensor data and determines whether an anomaly is present. It is prompted with a task-specific instruction and outputs both a binary decision (anomaly/no anomaly) and a textual explanation, as shown in Anomaly Detection Prompt. If an anomaly is detected, the relevant data is passed to the second LLM, which is responsible for classifying the fault into one of three types. 
This fault-classification LLM is provided with prior fault descriptions, including how each fault typically manifests in the sensor data (e.g., gradual increase in temperature, sudden pressure drop), embedded directly into the prompt as contextual information, as shown in Fault Detection Prompt.

This part of the study addresses two key research questions:  
\textbf{Q1.} Are LLMs inherently capable of detecting anomalies from sensor data across different data representations?  
\textbf{Q2.} Does the architecture of the LLM system influence fault classification performance?

To answer these questions, we generated synthetic time-series data spanning 10 days, sampled at 1-hour intervals. The dataset includes instances of three fault types: refrigerant leaks, compressor faults, and filter blockages, with controlled overlaps in their occurrence. Fault onset patterns were varied using a combination of three distinct onset profiles to evaluate the robustness of different system architectures. 

\begin{figure}[h]
    \centering
    \includegraphics[width=0.98\linewidth]{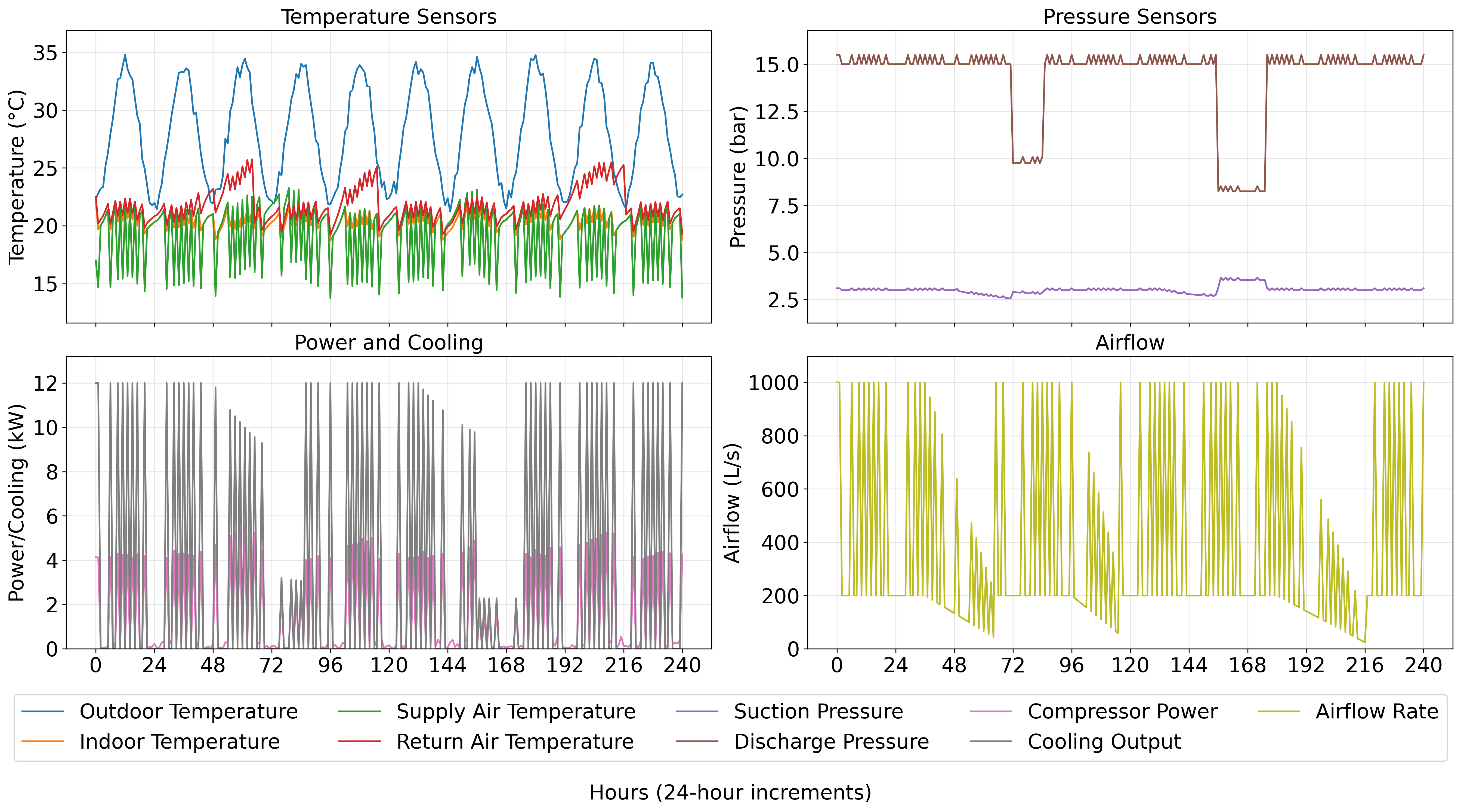}
    \caption{Visualization of simulated data with faults.}
    \label{fig:sensor_viz}
\end{figure}

\begin{figure}[h]
    \centering
    \includegraphics[width=0.5\linewidth]{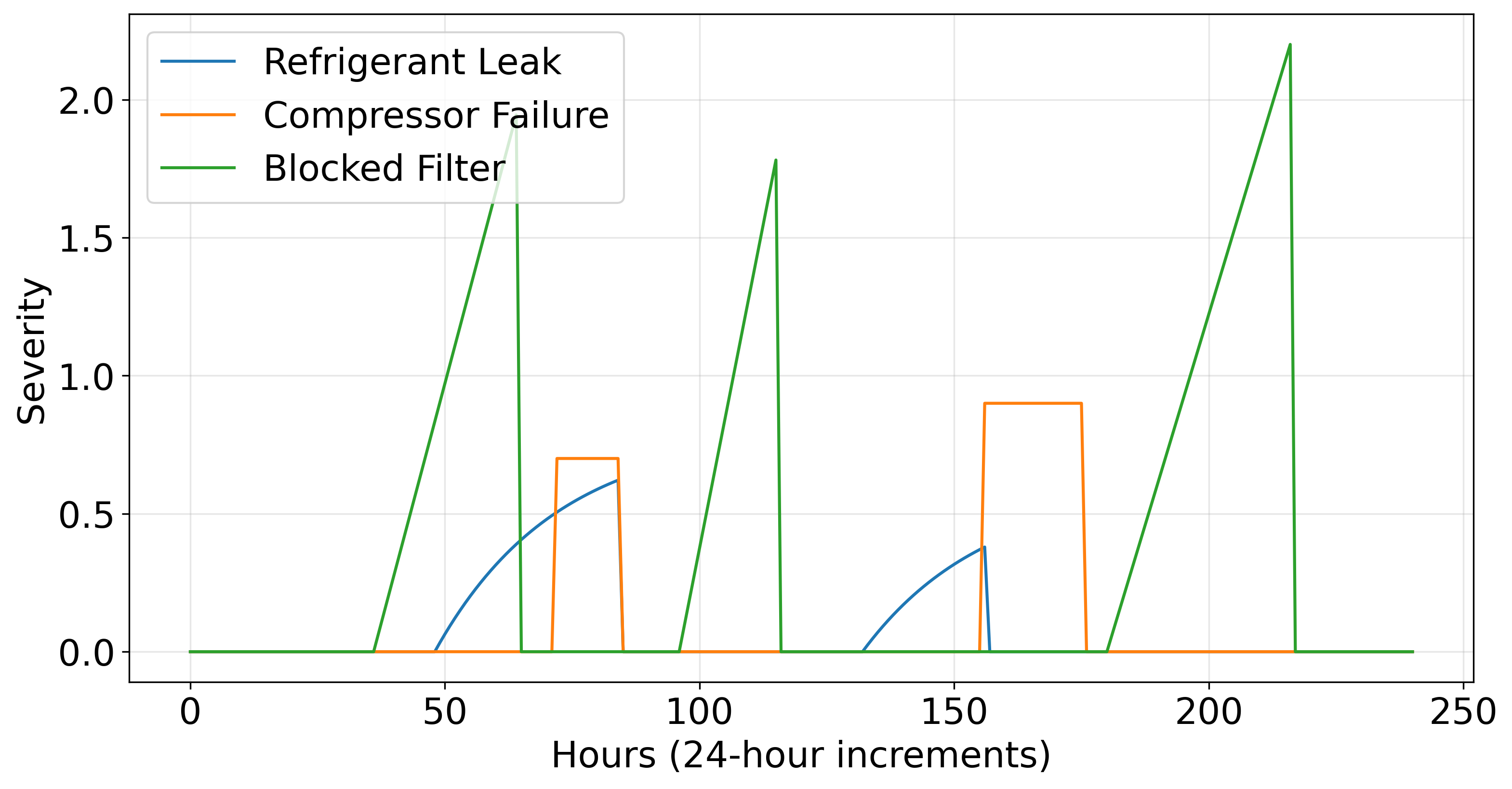}
    \caption{Visualization of fault onset profiles used to generate data in Fig~\ref{fig:sensor_viz}.}
    \label{fig:sensor_fault}
\end{figure}

Thee fault profiles and the corresponding sensor data are shown in Figure~\ref{fig:sensor_fault} and~\ref{fig:sensor_viz}. In all experiments, the LLM systems are given a sliding window of historical data (e.g., past 24 hours) and tasked with evaluating whether an anomaly, and subsequently a fault, has occurred in the most recent hour. The stride between evaluations is fixed at 1 hour, ensuring that detection resolution matches the sampling frequency.

\begin{tcolorbox}[promptbox, title=Anomaly Detection Prompt]
    \small
    You are an expert HVAC monitoring system. Your job is to analyze sensor data from HVAC systems
    to detect potential anomalies.\\

    When analyzing sensor data:\\
    - Look for unusual patterns or trends in the data\\
    - Consider relationships between different sensors\\
    
    Common anomaly patterns:\\
    - Sudden spikes or drops in readings\\
    - Unusual patterns in sensor relationships\\
    - Values outside normal operating ranges\\
    - Inconsistent behavior between related sensors\\

    Analyze the following HVAC sensor data to determine if the latest hour of data is anomalous with respect to the previous hours of data. \\

    Sensor data:\\
    \{Raw sensor data\}\\
    Statistics:\\
    \{Sensor data statistics\}\\
    Reference sensor data: \\
    \{Normal operational data - Raw data\}\\
    Reference sensor statistics \\
    \{Normal operational data - Statistics\}\\

    Is there evidence of any anomalies? Provide your analysis, including: \\
    - Concise, key observations from the sensor data \\
    - Whether you believe an anomaly is present (yes/no) \\
    - If an anomaly is present, provide a concise explanation for your conclusion

\label{anomaly_prompt}
\end{tcolorbox}

\begin{tcolorbox}[promptbox, title=Fault Detection Prompt, breakable]
    \small
    You are an expert HVAC fault classification system. Your job is to analyze sensor data and 
    classify the type of fault present in the system.\\

    When analyzing sensor data:\\
    - Look for unusual patterns or trends in the data\\
    - Consider relationships between different sensors\\

    Common fault patterns:\\
    - Refrigerant leak: Reduced suction pressure, reduced cooling output, increased compressor power and increased supply air temperature\\
    - Compressor failure: Reduced compressor power, reduced cooling output, increased suction pressure and decreased discharge pressure\\
    - Blocked filter: Reduced airflow rate, increased return air temperature and increased compressor power\\

    Analyze the following HVAC sensor data to classify the most probable type of faults present in the last hour of data with respect to the previous hours of data.\\

    Sensor data:\\
    \{Raw sensor data\}\\
    Statistics:\\
    \{Sensor data statistics\}\\

    Classify the type of fault present. For each fault type, indicate whether it is present (true/false):\\
    - Refrigerant leak\\
    - Compressor failure\\
    - Blocked filter\\
    Provide a concise, brief explanation for your classification
\label{fault_prompt}
\end{tcolorbox}

\subsection{Anomaly Detection and Fault Classification}

To answer \textbf{Q1}, we assess the LLM’s anomaly detection capabilities across the following experimental dimensions:

\begin{itemize}

    \item \textbf{Data representation:} We compare two input formats: 
    \begin{itemize}
        \item \textit{Raw data}, represented as a table (timestamps $\times$ sensor values).
        \item \textit{Descriptive statistics}, including min, max, mean, standard deviation, median, 25th and 75th percentiles, and trend (general direction over time). 
    \end{itemize}
    This comparison helps determine whether LLMs can natively infer statistics from raw data or benefit from explicit summary statistics.

    \item \textbf{ Use of reference data:} In some configurations, agents are provided with a reference sample of ordinary (non-faulty) operational data. This enables us to assess whether comparative reasoning (i.e., anomaly vs baseline) enhances the agent’s performance.
    
    \item \textbf{ Historical context window size:} We vary the window length of past data to assess whether a longer temporal context improves performance.
    
    \item \textbf{LLM model variant:} We evaluate performance across model scales, including \texttt{GPT-4.1-nano} and \texttt{GPT-4o}, to study how model size impacts detection and classification accuracy.
    
\end{itemize}

To answer \textbf{Q2}, we compare two system architectures:
\begin{itemize}
    \item \textbf{Centralized:} A single agent performs both anomaly detection and fault classification.
    \item \textbf{Decentralized:} Multiple agents, each dedicated to detecting a specific fault type independently.
\end{itemize}

This comparison allows us to evaluate the trade-offs between specialization and generalization. The single-LLM system benefits from global context and shared knowledge across all fault types, but it may suffer from task interference or limited capacity in recognizing multiple fault dynamics simultaneously. Conversely, the multi-LLM setup allows each LLM to focus on a narrower, fault-specific detection problem, potentially improving sensitivity and interpretability for individual fault types.

Through this experiment, we aim to determine whether fault-specific specialization enhances detection and classification accuracy, or if a single generalized LLM can perform competitively across a range of fault scenarios. This also has implications for real-world deployment: single-LLM systems may be easier to scale and maintain, while multi-LLM systems could offer higher robustness in environments where fault types have highly distinct signatures. Note that for fair comparison, all our experiments uses API calls to commercial LLMs with default values for inference-time hyperparameters, such as top-p and temperature values. 

\subsubsection{Statistical rule baseline}
As a baseline, we developed a simple rule-based method to compare against the LLM-agent. For anomaly detection, a time window is evaluated for deviations from the statistical patterns observed in historical normal operation data. For each sensor, we compute the minimum, maximum, mean $\mu$, and standard deviation $\sigma$ based on the normal operational data. A window is flagged as anomalous if it violates any of the following criteria: if the minimum or maximum value within the window falls outside the historical bounds, i.e., $\min(x_{\text{window}}) < \min(x_{\text{normal}})$ or $\max(x_{\text{window}}) > \max(x_{\text{normal}})$; if the variability in the window is unusually high such that $\sigma_{\text{window}} > \sigma_{\text{normal}}$; or if there is a sudden trend shift, measured by splitting the window in half and comparing the means of each half, where $|\mu_2 - \mu_1| > 2\sigma_{\text{normal}}$ indicates a significant change.

For fault classification, we rely on domain knowledge to define rule-based conditions that capture characteristic combinations of sensor behaviors, with thresholds derived from each sensor’s mean and standard deviation under normal conditions. A $F_{leak}$ is inferred when suction pressure and cooling output fall at least one standard deviation below their respective means, while compressor power and supply air temperature rise above the same amount. In contrast, a $F_{comp}$ is identified by a drop in compressor power and cooling output, both one standard deviation below their means, accompanied by elevated suction pressure and reduced discharge pressure, each deviating from the mean by at least one standard deviation in opposite directions. A $F_{filter}$ is detected when the airflow rate is drops relative to its mean, while return air temperature and compressor power simultaneously exceed their normal levels by a similar margin of one standard deviation above the mean.

While this may not represent the most sophisticated baseline, and more advanced statistical or machine learning methods could yield better performance, we consider it a valuable and appropriate comparison due to its simplicity and ease of implementation.

\subsection{Continual Learning}
\label{subsec:continual_learning}
In the second part of our study, we evaluate the LLM system’s ability to adapt over time in an online setting using in-context learning principles~\cite{dong2022survey}. We simulate a human-in-the-loop feedback process where an expert reviews the system’s predictions, provides corrected labels, and these corrections are incorporated into subsequent prompts. This approach emulates continual learning through prompt-based feedback, without altering the model’s internal weights.

This process can also be viewed through the lens of few-shot prompting in a temporal context: at each evaluation cycle, the LLM receives a prompt containing labeled examples from previous cycles. This design tests whether incremental prompting can improve fault classification performance over time.

To isolate the effects of the continual learning mechanism, we simulated a dedicated dataset with a single fault type, filter blockage, that recurs periodically over 20 days, and sampled hourly. Fault instances follow randomized onset profiles (e.g., linear progression with varying severity), as shown in Figure~\ref{fig:sub_a}. Each fault event lasts no more than one day and recurs periodically at first, with a long fault-free interval inserted towards the end. 

We define a learning cycle as one evaluation-feedback-update loop. In each cycle, performance is measured, and the labeled examples from the prior cycle is appended to the next prompt. This setup enables a systematic assessment of the LLM’s ability to leverage historical feedback to improve anomaly detection accuracy across successive cycles. 

\section{Results and Discussion}

\subsection{Anomaly detection results}

\begin{table*}[h]
\small
\centering
\caption{Performance of GPT-4.1-nano Across Various Input Conditions}
\begin{tabular}{|c|c|c|c|c|c|c|}
\hline
\textbf{Model} & \textbf{Data Representation} & \textbf{Reference Data Representation} & \textbf{Precision} & \textbf{Recall} & \textbf{F1-Score} & \textbf{Accuracy} \\
\hline
\multirow{12}{*}{GPT-4.1-nano}
 & Descriptive statistics & Both & 0.73 & 0.99 & 0.84 & 0.73 \\
 & Descriptive statistics & Descriptive statistics & 0.73 & 0.99 & 0.84 & 0.73 \\
 & Descriptive statistics & None & 0.73 & 0.99 & 0.84 & 0.72 \\
 & Descriptive statistics & Raw Data & 0.73 & 0.99 & 0.84 & 0.72 \\
 & Both & Descriptive statistics & 0.73 & 0.95 & 0.83 & 0.71 \\
 & Both & None & 0.73 & 0.93 & 0.82 & 0.70 \\
 & Both & Both & 0.73 & 0.93 & 0.82 & 0.69 \\
 & Raw Data & None & 0.73 & 0.91 & 0.81 & 0.69 \\
 & Raw Data & Descriptive statistics & 0.72 & 0.93 & 0.81& 0.68 \\
 & Raw Data & Both & 0.72 & 0.91 & 0.80 & 0.68 \\
 & Both & Raw Data & 0.72 & 0.88 & 0.79 & 0.66 \\
 & Raw Data & Raw Data & 0.75 & 0.84 & 0.79 & 0.67 \\
\hline
Rule-based & NA & NA & 0.73 & 1.00 & 0.85 & 0.73\\
\hline
\end{tabular}
\label{tab:gpt41_performance}
\end{table*}

Table~\ref{tab:gpt41_performance} lists the performance of the LLM system in anomaly detection, sorted in descending order of F1-score—the harmonic mean of precision and recall. Several observations can be made from these results. First, in terms of input data representation, when the LLM system is provided with either raw sensor data, descriptive statistics, or both, the top-performing configurations are generally those that use descriptive statistics. These are followed by setups that combine both types of input, and finally by those using only raw data. This suggests that summarizing raw numerical values into key descriptors is beneficial for anomaly detection. This trend is further supported by the fact that configurations using both representations tend to fall in the middle of the performance ranking, while those relying solely on raw data consistently rank lowest.

With respect to the inclusion of reference data from normal operations, there is no consistent pattern indicating whether its presence, or the representation in which it is included, has a significant effect. This suggests that reference data, in this context, may not be particularly beneficial. Another consistent pattern is that recall is higher than precision, indicating that the model tends to produce more positive predictions, potentially resulting in more false positives. Precision itself remains relatively stable across all configurations, ranging narrowly between 0.72 and 0.75, which implies that the variation in F1-score is primarily driven by differences in recall.

It is also worth noting that accuracy does not align well with F1-score in this setting. For example, the highest-performing configuration based on F1-score (input data in the form of descriptive statistics with both representations of reference data) achieves an F1 of 0.84 but an accuracy of only 0.73, while one of the lowest-performing setups in terms of F1 (raw input data with raw reference data) still achieves an accuracy of 0.67. This discrepancy arises from class imbalance, as shown in the fault distribution in Figure~\ref{fig:sensor_fault}, which leads to artificially inflated accuracy when the model overpredicts the dominant class. As such, F1-score offers a more reliable measure of anomaly detection performance in this task. For comparison, we also include the performance of a rule-based statistical baseline. The results show that with the appropriate data representation, the LLM system can approach the baseline’s performance. However, under suboptimal input conditions, the LLM system tends to lag behind.

To illustrate the model’s behavior more concretely, we anecdotal examples of the LLM system’s predictions, covering both true positives and false positives. As seen in the key observations generated by the LLM system, it is generally effective at identifying and describing numerical trends. However, a common failure point lies in its lack of causal grounding and domain-specific operational knowledge, which often results in false positives. For instance, the LLM system may flag an airflow spike as anomalous purely because it is statistically extreme, even when provided with reference data, whereas a domain expert would recognize it as expected during high-demand periods, such as peak cooling hours in summer.

\begin{figure}[t]
\noindent
\begin{tcolorbox}[promptbox, title=Output: True positive prediction]
    \small
    \textbf{Key observation:} 
    The latest hour shows a significant increase in outdoor temperature to 21.98\si{\celsius}, a rise in indoor temperature to 20.33\si{\celsius}, and a spike in compressor power to 0.23 kW, with the airflow rate returning to 200, which is typical for non-operational or low operation states.\\

    \textbf{Predicted anomaly:} 
    True\\

    \textbf{Explanation:}
    The sudden spike in outdoor and indoor temperatures, combined with the increase in compressor power and the low airflow rate, suggests an abnormal change in system behavior, possibly indicating a fault or external environmental influence affecting the HVAC system's normal operation.\\

    \textbf{Ground truth:}
    True
\label{sample_output_tp}
\end{tcolorbox}
\end{figure}

This points to a tendency toward statistical over-reliance, particularly when the LLM is making predictions without sufficient system-level context. In such cases, the model may overgeneralize from statistical norms and fail to recognize contextually valid exceptions. Nonetheless, the LLM's ability to provide detailed explanations for its predictions is a valuable feature. These explanations offer insight into the model’s reasoning process and can serve as a useful tool in future work. By systematically analyzing them, operators can identify recurring reasoning patterns and use these insights to design better prompts or input structures that incorporate domain knowledge, ultimately improving anomaly detection accuracy and reducing misclassifications.

Next, the results in Table~\ref{tab:model_window_comparison} examines the effects of both model size and input window size. As the results show, varying the window size does impact performance, with a window size of 36 emerging as the most effective in our setting. Increasing the window size to 48 yields a slight drop in F1-score for both models, and decreasing it to 24 results in a more noticeable decline. We believe the optimal window size depends in part on the nature of the faults. Larger windows may overwhelm the model with too much context or dilute fault-relevant patterns, while smaller windows may fail to provide enough context for the LLM to identify the anomaly. Although we cannot conclude a single ideal window size across all scenarios, these results suggest that tuning this parameter is important for maximizing performance.

\begin{figure}[t]
\noindent
\begin{tcolorbox}[promptbox, title=Output: False positive prediction]
    \small
        \textbf{Key observation:} 
    The latest hour shows a significant spike in airflow rate to 1000, which is well above the 75th percentile (319.44) and the maximum of previous data (1000). Additionally, compressor power and cooling output are also elevated compared to their typical ranges, with compressor power reaching 4.43 and cooling output at 9.29, both higher than their median values.\\

    \textbf{Predicted anomaly:} 
    True\\

    \textbf{Explanation:}
    The sudden and extreme increase in airflow rate to 1000, which is substantially above the typical maximum and the interquartile range, indicates an abnormal behavior. The concurrent spikes in compressor power and cooling output further support the presence of an anomaly, likely due to a sensor malfunction or an operational fault.\\

    \textbf{Ground truth:}
    False
\label{sample_output_fp}
\end{tcolorbox}
\end{figure}

\begin{table*}[h]
\small
\centering
\caption{Anomaly Detection Performance Comparison Between Model and Window Size (Input Source: Statistics, Input Normal Example: Both)}
\begin{tabular}{|c|c|c|c|c|c|}
\hline
\textbf{Model} & \textbf{Window Size} & \textbf{Precision} & \textbf{Recall} & \textbf{F1-Score} & \textbf{Accuracy} \\
\hline
\multirow{3}{*}{GPT-4o}
 & 36 & 0.73 & 0.99 & 0.84 & 0.73 \\
 & 48 & 0.72 & 1.00 & 0.84 & 0.72 \\
 & 24 & 0.69 & 0.99 & 0.82 & 0.69 \\
\hline
\multirow{3}{*}{GPT-4.1-nano}
 & 36 & 0.73 & 0.99 & 0.84 & 0.73\\
 & 48 & 0.71 & 0.97 & 0.82 & 0.69 \\
 & 24 & 0.69 & 0.98 & 0.81 & 0.68 \\
\hline
Rule-based & 36 & 0.73 & 1.00 & 0.85 & 0.73 \\
\hline
\end{tabular}
\label{tab:model_window_comparison}
\end{table*}

We also observe that model size has a modest but consistent effect. Switching \texttt{GPT-4.1-nano} to \texttt{GPT-4o} leads to a small improvement across nearly all metrics, especially recall and F1-score. While the gains are not large, they are consistent, suggesting that model capacity can help the LLM make more confident or accurate anomaly predictions, particularly when more context is available.

Finally, for reference, we include the rule-based baseline, which again slightly outperforms all LLM variants in terms of F1-score. However, the best-performing LLM configuration (\texttt{GPT-4o} with a window size of 36) comes very close, indicating that with appropriate tuning, LLMs can approach traditional baselines in this domain.

\subsection{Fault classification results}

Building on these findings, Table~\ref{tab:fault_classification_performance} explores fault classification performance across two different LLM models, \texttt{GPT-4o} and \texttt{GPT-4.1-nano}, under varying LLM system architectures and reference data representations, given both raw data and descriptive statistics as inputs. The primary metric again is F1-score, reflecting the balance between precision and recall in this imbalanced classification task. The results show that the multi-LLM architecture, where three specialized LLMs, each focusing on one fault type, outperforms the single-LLM approach, which relies on a single LLM to detect all faults simultaneously. For \texttt{GPT-4o}, the model achieves an F1-score of around 0.59 with multi-LLM architectures, which marginally but consistently surpass those of the single-LLM architecture that reach an F1-score of at most 0.58. Meanwhile, \texttt{GPT-4.1-nano} performs substantially worse across the board, with F1-scores near 0.43 in multi-LLM setups and below 0.36 in single-LLM conditions.

Similar to the anomaly detection results, the representation or presence of reference data appears to have limited impact on classification performance, reinforcing the idea that including reference data representation is not beneficial in this context. A closer look at recall and precision reveals that multi-LLM architectures deliver very high recall, above 0.90 for both models, indicating strong sensitivity to fault presence, though precision remains low, especially for the smaller model. We also note that single-LLM architectures trade recall for modestly improved precision, which aligns with their comparatively lower F1-scores.

The rule-based baseline, despite achieving the highest accuracy of 0.71 by predominantly predicting the dominant negative class, fails entirely at fault classification, with zero precision, recall, and F1-score. This contrast underscores the difficulty of the task and highlights that while LLM-based models do not yet fully resolve these challenges, especially under single-LLM settings or smaller model conditions, they may provide valuable and more nuanced fault detection capabilities beyond simple heuristics. Overall, these results suggest that specialized multi-LLM architectures improve fault classification effectiveness, particularly for more capable models like \texttt{GPT-4o}, while reference data inclusion remains secondary.

\begin{table*}[h]
\small
\centering
\caption{Fault Classification Performance Across Model and System Classification Architecture}
\begin{tabular}{|c|c|c|c|c|c|c|}
\hline
\textbf{Model} & \textbf{Reference Data Representation} & \textbf{Architecture} & \textbf{Precision} & \textbf{Recall} & \textbf{F1-Score} & \textbf{Accuracy} \\
\hline
\multirow{8}{*}{GPT-4o}
 & Both & Decentralized & 0.48 & 0.94 & 0.59 & 0.56 \\
 & Raw Data & Decentralized & 0.46 & 0.95 & 0.59 & 0.55 \\
 & Descriptive statistics & Decentralized & 0.47 & 0.95 & 0.59 & 0.55 \\
 & None & Decentralized & 0.46 & 0.94 & 0.59 & 0.56 \\
 & None & Centralized & 0.49 & 0.76 & 0.58 & 0.69 \\
 & Raw Data & Centralized & 0.49 & 0.73 & 0.58 & 0.68 \\
 & Descriptive statistics & Centralized & 0.48 & 0.71 & 0.57 & 0.68 \\
 & Both & Centralized & 0.47 & 0.74 & 0.56 & 0.67 \\
\hline
\multirow{8}{*}{GPT-4.1-nano}
 & None & Decentralized & 0.29 & 0.95 & 0.44 & 0.29 \\
 & Descriptive statistics & Decentralized & 0.29 & 0.96 & 0.43 & 0.29 \\
 & Raw Data & Decentralized & 0.28 & 0.97 & 0.43 & 0.28 \\
 & Both & Decentralized & 0.28 & 0.96 & 0.44 & 0.28 \\
 & Descriptive statistics & Centralized & 0.37 & 0.36 & 0.36 & 0.66 \\
 & None & Centralized & 0.37 & 0.35 & 0.36 & 0.67 \\
 & Raw Data & Centralized & 0.35 & 0.36 & 0.35 & 0.67 \\
 & Both & Centralized & 0.33 & 0.32 & 0.32 & 0.64 \\
\hline
Rule-based & NA & NA & 0 & 0 & 0 & 0.71 \\
\hline
\end{tabular}
\label{tab:fault_classification_performance}
\end{table*}

\subsection{Continual learning results}


\begin{figure}[h]
    \centering

    \begin{subfigure}{\linewidth}
        \centering
        \includegraphics[width=\linewidth]{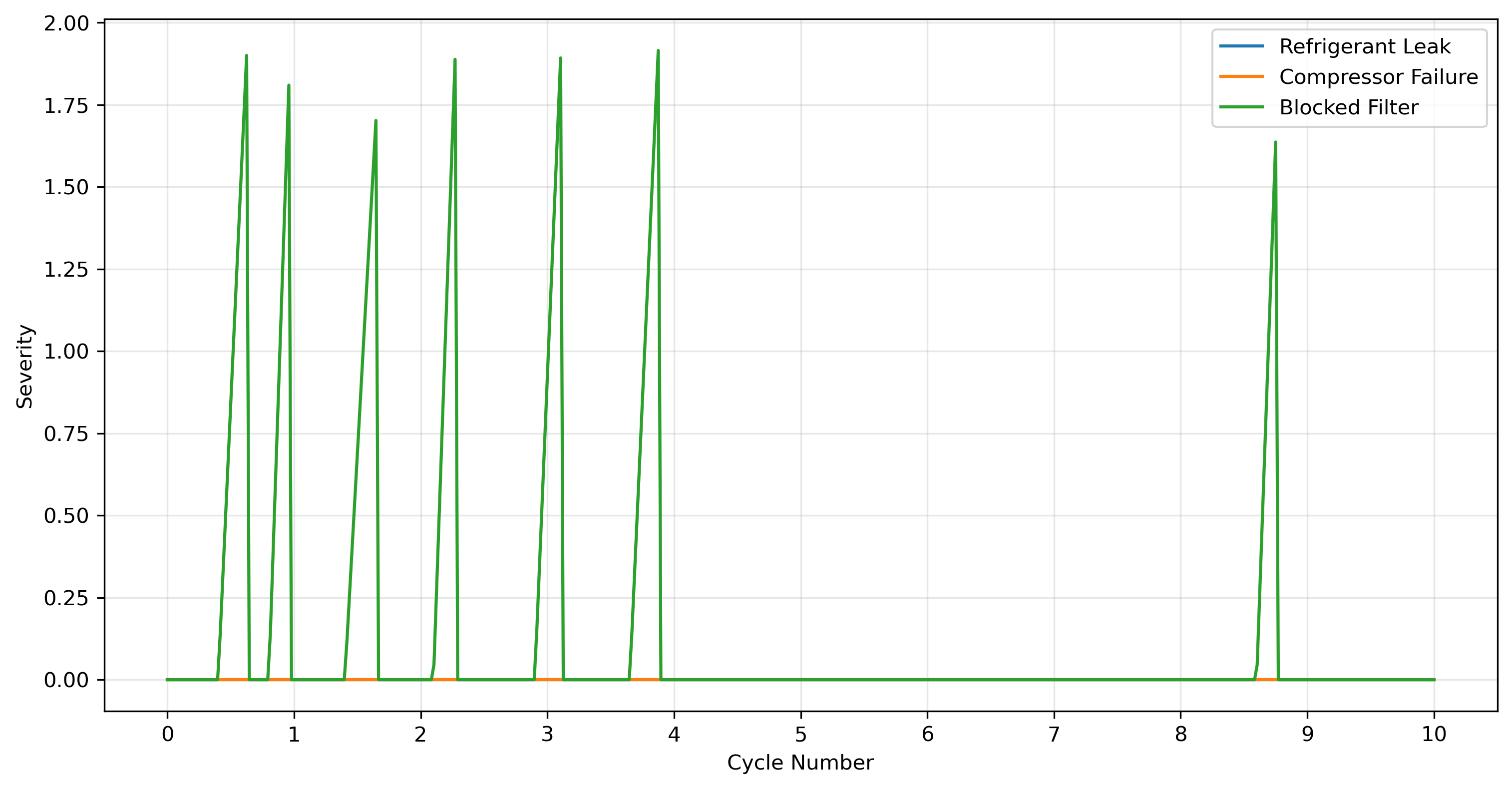}
        \caption{}
        \label{fig:sub_a}
    \end{subfigure}

    \begin{subfigure}{\linewidth}
        \centering
        \includegraphics[width=\linewidth]{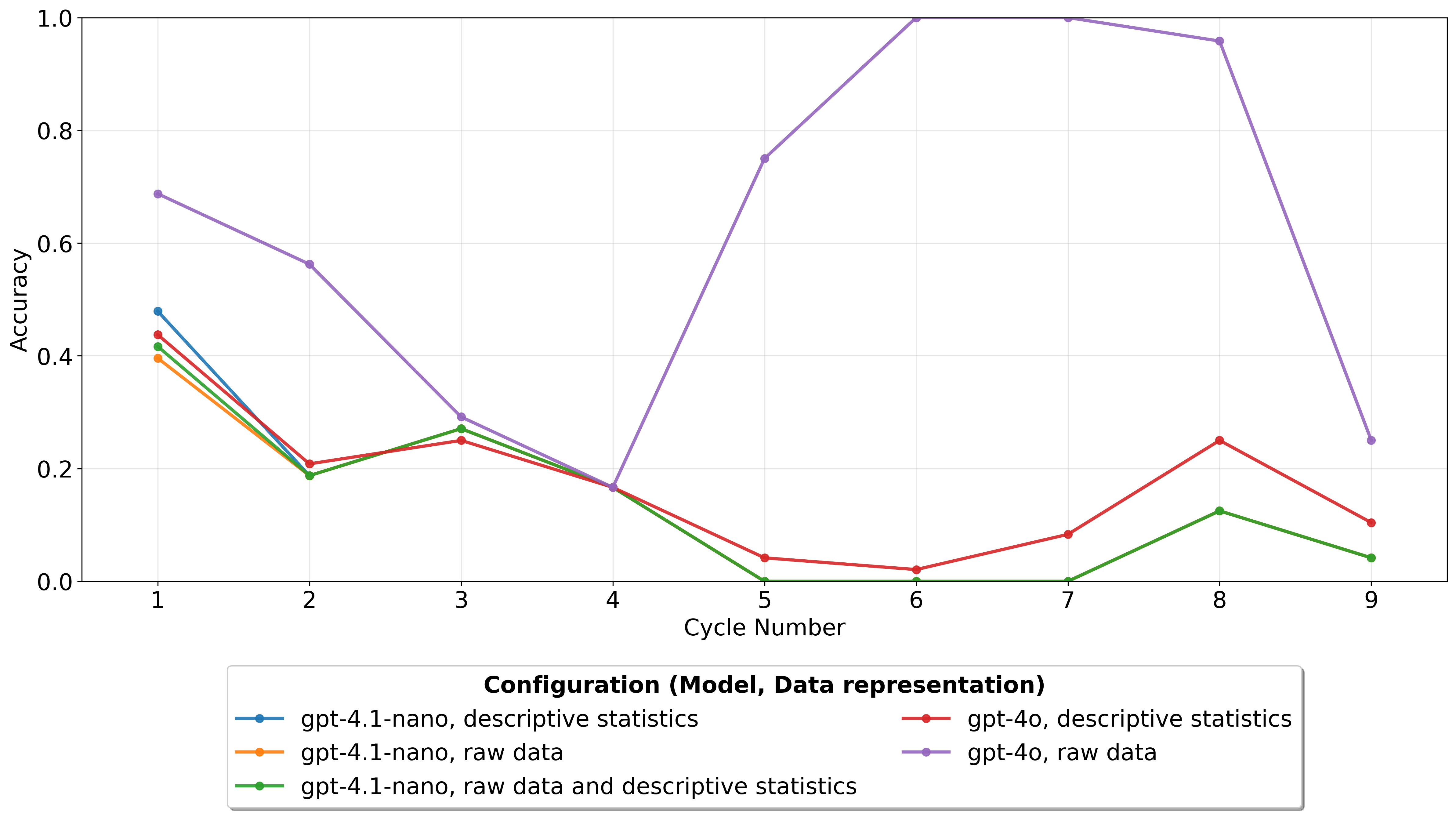}
        \caption{}
        \label{fig:sub_b}
    \end{subfigure}

    \caption{(a) Visualization of the fault onset profile used to generate data for the continual learning experiments. (b) Performance (anomaly detection accuracy) of model configurations across continual learning cycles.}
    \label{fig:combined}
\end{figure}

In this section, we delve deeper into the experiments investigating the capability of LLMs to learn continually in an online setting. Specifically, we examine whether these LLMs can improve their performance on the fly when provided with feedback in the form of their own previous predictions alongside the corresponding ground truth, using the data described in Subsection~\ref{subsec:continual_learning}. Figure~\ref{fig:sub_a} illustrates the onset of the fault profile, while Figure~\ref{fig:sub_b} shows the progression of accuracy achieved by the anomaly detection LLM over multiple cycles. In this experiment, we use accuracy as the primary evaluation metric rather than F1 score, as there are extended periods in the dataset where no faults occur. During such periods, both precision and recall, and consequently the F1 score, drop to zero, providing limited insight.

Intuitively, as historical information accumulates, we would expect detection performance to improve. The combination of past predictions and their ground truth labels could serve as in-context learning examples, enabling the LLM to refine its internal representation of faults. However, contrary to this expectation, our results (Figure~\ref{fig:sub_b}) show that most LLMs do not exhibit effective continual learning. In fact, accuracy declines in the early cycles and remains consistently low in later ones. This trend suggests a growing confusion in the presence of repeated fault events. From cycle 4 onward, when no faults are present, the near-zero accuracy indicates a persistent bias toward predicting faults, resulting in a high rate of false positives.

Moreover, most LLMs fail to correct this bias during fault-free intervals, continuing to predict faults even when none exist. An exception is observed for \texttt{GPT-4o} that is configured to use only raw sensor data. Although its accuracy also drops in the initial cycles, it begins to recover starting in cycle 4, coinciding with the absence of faults, and gradually returns to 100\% accuracy. This pattern may suggest that the LLM is incorporating feedback over time in a few-shot learning manner. However, this apparent improvement is brought into question by the LLM's failure to detect the final fault event, during which it incorrectly predicts a normal operation.

These findings indicate that \texttt{GPT-4o} model relying on raw data may be heavily influenced by trends in historical input, potentially at the expense of sensitivity to current sensor readings. The improvement in accuracy may reflect a form of input pattern matching rather than true learning from feedback. This supports our earlier hypothesis that while LLMs using raw data can detect temporal patterns, they lack sufficient grounding in system behavior to perform reliable anomaly detection across changing conditions.

\subsection{Discussion}

In this study, we evaluated two large language models, \texttt{GPT-4.1-nano} and \texttt{GPT-4o}, for fault detection in a simulated HVAC system. While the models demonstrated strong potential, many LLMs with more explicit reasoning capabilities remain unexplored. These alternatives may offer improved performance but often involve higher computational costs and inference latency, which are important trade-offs for real-time applications. In parallel, an important comparison for future work lies in evaluating LLM-based approaches against advanced non-LLM methods, such as physics-informed models, rule-based expert systems, or traditional machine learning pipelines, which may offer different tradeoffs of efficiency, usability,  interpretability, robustness, and modularity. For example, data-driven approaches may be more generalizable if they're trained on a large amount of historical data while rule-based expert systems may be more interpretable and reliable as they're grounded in domain knowledge.

Additionally, our prompt design in this work focused on producing natural, user-like queries to align with practical usage patterns. Given the vast and non-exhaustive nature of the prompt space, we limited our scope to a single, interpretable prompt per setting. More sophisticated prompting strategies, including few-shot prompting, tool use, or chain-of-thought reasoning, represent valuable directions for future exploration. It is also worth noting that prompt complexity itself may act as a confounding factor when comparing single-LLM versus multi-LLM architectures, since richer reasoning styles (e.g., CoT) may yield disproportionate gains or penalties depending on system organization.

Similarly, while we incorporated a mechanism for continual learning, more work is needed to study and develop a robust adaptation mechanism over time as system behavior evolves. With the pace of ongoing improvements in LLMs, future models may significantly impact both reasoning capability and deployment feasibility as well. While general-purpose LLMs offer flexibility and ease of interaction, domain-specific models or hybrid approaches could yield greater transparency and alignment with engineering practice. Exploring such domain-specific solutions, either through fine-tuned models or tightly coupled physics–data hybrids, represents another valuable research path.

Another adjacent direction for future work is to use the HVAC simulator developed in this study to evaluate the ability of LLMs to distinguish between true system faults and sensor drift. This is a particularly challenging diagnostic task that is common in real-world industrial systems. The simulator’s ability to generate controlled, configurable fault and drift scenarios offers a useful platform to investigate how well LLMs can reason about subtle differences in temporal patterns and fault signatures. Advancing this capability would be an important step toward building LLM systems that are both accurate and trustworthy in practical monitoring applications. Finally, connecting these findings more explicitly to real-world industrial deployment remains an open challenge. Issues such as system integration, safety assurance, regulatory compliance, and operator trust needs to be addressed to ensure that LLM-driven monitoring systems transition from controlled studies into reliable, field-ready tools.

\section{Conclusion}

LLM-based systems show promise for fault detection in sensor-driven industrial environments, offering advantages in usability, explainability, and accessibility over traditional methods. In our experiments, most LLMs either under perform or match the performance of simple rule-based statistical baselines in anomaly detection. For fault classification, rule-based methods achieve higher accuracy, while LLMs tend to perform better in precision and recall, highlighting a nuanced trade-off between approaches. However, these differences are often small and may not justify the higher computational cost of LLMs. Initial results from continual learning experiments show minimal improvement, indicating that further development is needed to realize adaptive capabilities. Future work will focus on improving continual learning effectiveness, exploring more advanced reasoning-oriented LLMs, and designing hybrid systems that combine rule-based logic with LLM-driven analysis to leverage the strengths of both.

\section*{Nomenclature}
\begin{tabular}{ l  p{5.5cm} }
        $T_{in}$			& Indoor temperature \\
	$T_{amb}$			& Ambient (outdoor) temperature \\
	$T_{mean}$		& Mean ambient temperature \\
	$T_{supply}$		& Supply air temperature \\
	$T_{return}$		& Return air temperature \\
	$Q_{cool}$		& Cooling output \\
	$Q_{nom}$		& Nominal cooling capacity \\
	$Q_{air}$       & Airflow rate \\
	$P_{suct}$		& Suction pressure of refrigerant \\
	$P_{disc}$		& Discharge pressure of refrigerant \\
	$P_0$			& Base refrigerant pressure \\
	$P_{comp}$		& Compressor power consumption \\
	$P_{nom}$		& Nominal compressor power \\
	$\alpha$		& Thermal gain coefficient \\
	$\beta$			& Cooling effectiveness \\
	$\gamma_1$, $\gamma_2$ & Pressure scaling coefficients \\
	$\phi$			& Phase offset for ambient temperature \\
	$A$			    & Ambient temperature oscillation amplitude\\
	$\epsilon(t)$   & Ambient temperature noise, $\mathcal{N}(0, \sigma^2)$ \\
	$\eta(t)$		& Compressor power variation, $\mathcal{N}(0, \sigma_p^2)$ \\
	$f(t)$			& Fault onset function \\
	$F(t)$			& Time-varying fault impact \\
	$F_{leak}$		& Refrigerant leak fault indicator \\
	$F_{comp}$		& Compressor fault indicator \\
	$F_{filter}$		& Filter blockage fault indicator \\
	$S$			& Fault severity scalar \\
	$t_0, t_1$		& Fault ramp start and end times \\
	$\tau$			& Fault exponential time constant \\
\end{tabular}

\bibliography{references}  
\end{document}